\documentclass[10pt,journal]{IEEEtran}
\usepackage[square,sort,comma,numbers]{natbib}
\usepackage{times}
\usepackage{latexsym}
\usepackage{graphicx}
\usepackage{hyperref}
\usepackage{csquotes}
\usepackage{amsmath}
\usepackage{amsfonts}

\title{An Attention Mechanism for Neural Answer Selection Using a Combined Global and Local View}

\author{Yoram Bachrach, Andrej \v{Z}ukov-Gregori\v{c}, Sam Coope, Ed Tovell, \\
Bogdan Maksak, Jos\'{e} Rodriguez, Conan McMurtie, Mahyar Bordbar \\
    \texttt{\{yoram,andrej,sam,ed,bogdan,jose,conan,mario\}@digitalgenius.com}
}

\date{}

\hypersetup{draft}
\begin{document}

\maketitle

\begin{abstract}
We propose a new attention mechanism for neural based question answering, which depends on varying granularities of the input. 
Previous work focused on augmenting recurrent neural networks for question answering systems with simple attention mechanisms which are a function of the similarity between a question embedding and an answer embeddings across time. 
We extend this by making the attention mechanism dependent on a global embedding of the answer attained using a separate network. 
We evaluate our system on InsuranceQA, a large question answering dataset. Our model outperforms current state-of-the-art results on InsuranceQA. 
Further, we examine which sections of text our attention mechanism focuses on, and explore its performance across different parameter settings.

 \end{abstract}

\section{Introduction}
Question answering (QA) relates to the building of systems capable of automatically answering questions posed by humans in natural language. Various frameworks have been proposed for question answering, ranging from simple information-retrieval techniques for finding relevant knowledge articles or webpages, through methods for identifying the most relevant sentence in a text regarding a posed question, to methods for querying structured knowledge-bases or databases to produce an answer~\cite{burke1997question,voorhees1999trec,kwok2001scaling,hirschman2001natural,ravichandran2002learning}

A popular QA task is {\em answer selection}, where, given a question, the system must pick correct answers from a pool of candidate answers~\cite{xu2002trec,jijkoun2004answer,ko2007probabilistic,lee2009model,severyn2013automatic}.

Answer selection has many commercial applications. Virtual assistants such as Amazon Alexa and Google Assistant are designed to respond to natural language questions posed by users. In some cases such systems simply use a search engine to find relevant webpages; however, for many kinds of queries, such systems are capable of providing a concise specific answer to the posed question. 

Similarly, various AI companies are attempting to improve customer service by automatically replying to customer queries. One way to design such a system is to curate a dataset of historical questions posed by customers and the responses given to these queries by human customer service agents. Given a previously unobserved query, the system can then locate the best matching answer in the curated dataset. 

Answer selection is a difficult task, as typically there is a large number of possible answers which need to be examined. Furthermore, although in many cases the correct answer is lexically similar to the question, in other cases semantic similarities between words must be learned in order to find the correct answer~\cite{kolomiyets2011survey,allam2012question}. Additionally, many of the words in the answer may not be relevant to the question. 

Consider, for example, the following question answer pair:

\begin{displayquote}
\textbf{How do I freeze my account?}

Hello, hope you are having a great day. You can freeze your account by logging into our site and pressing the freeze account button. Let me know if you have any further questions regarding the management of your account with us. 
\end{displayquote}

\noindent Intuitively, the key section which identifies the above answer as correct is ``[...] you can freeze your account by [...]'', which represents a small fraction of the entire answer.

Earlier work on answer selection used various techniques, ranging from information retrieval methods~\cite{clarke2001exploiting} and machine learning methods relying on hand-crafted features~\cite{parsetreeManning,wang2007jeopardy}. Deep learning methods, which have recently shown great success in many domains including image classification and annotation~\cite{krizhevsky2012imagenet,zhou2014learning,lewenberg2016predicting}, multi-annotator data fusion~\cite{albarqouni2016aggnet,gaunt2016training}, NLP and conversational models~\cite{graves2013speech,bahdanau2014ntm,li2015diversity,kandasamy2017batch,shao2017generating} and speech recognition~\cite{graves2013speech,albarqouni2016aggnet}, have also been successfully applied to question answering~\cite{fengCNN}. Current state-of-the-art methods use recurrent neural network (RNN) architectures which incorporate attention mechanisms~\cite{tan2016}.  These allow such models to better focus on relevant sections of the input~\cite{bahdanau2014ntm}.

{\bf Our contribution: } We propose a new architecture for question answering. Our high-level approach is similar to recently proposed QA systems~\cite{fengCNN,tan2016}, but we augment this design with a more sophisticated attention mechanism, combining the {\em local} information in a specific part of the answer with a {\em global} representation of the entire question and answer. 

We evaluate the performance of our model using the recently released {\em InsuranceQA dataset}~\cite{fengCNN}, a large open dataset for answer selection comprised of insurance related questions such as: ``what can you claim on Medicare?''. \footnote{As opposed to other QA tasks such as answers extraction or machine text comprehension and reasoning~\cite{weston2015towards,rajpurkar2016squad}, the InsuranceQA dataset questions do not generally require logical reasoning.}

We beat state-of-the-art approaches ~\cite{fengCNN,tan2016}, and achieve good performance even when using a relatively small network. 



\section{Previous Work}

Answer selection systems can be evaluated using various datasets consisting of questions and answers. Early answer selection models were commonly evaluated against the QASent dataset \cite{wang2007jeopardy}; however, this dataset is very small and thus less similar to real-world applications. Further, its candidate answer pools are created by finding sentences with at least one similar (non-stopword) word as compared to the question, which may create a bias in the dataset. 

Wiki-QA~\cite{yang2015wikiqa} is a dataset that contains several orders of magnitude more examples than QASent, where the candidate answer pools were created from the sentences in the relevant Wikipedia page for a question, reducing the amount of keyword bias in the dataset compared to QASent. 

Our analysis is based on the InsuranceQA~\cite{fengCNN} dataset, which is much larger, and similar to real-world QA applications. The answers in InsuranceQA are relatively long (see details in Section~\ref{sec:setup}), so the candidate answers are likely to contain content that does not relate directly to the question; thus, a good QA model for InsuranceQA must be capable of identifying the most important words in a candidate answer.

Early work on answer selection was based on finding the semantic similarity between question and answer parse trees using hand-crafted features \cite{parsetreeManning, wang2007jeopardy}. Often, lexical databases such as WordNet were used to augment such models \cite{ChangWordnet}. Not only did these models suffer from using hand-crafted features, those using lexical databases were also often language-dependent. 

Recent attempts at answer selection aim to map questions and candidate answers into n-dimensional vectors, and use a vector similarity measure such as cosine similarity to judge a candidate answer's affinity to a question. In other words, the similarity between a question and a candidate is high if the candidate answers the question well, low if the candidate is not a good match for the question. 

Such models are similar to Siamese models, a good review of which can be found in Muller et al's paper~\cite{mueller2016siamese}. Feng et al.~\cite{fengCNN} propose using convolutional neural networks to vectorize both questions and answers before comparing them using cosine similarity. Similarly, Tan et al.~\cite{tan2016} use a recurrent neural network 
to vectorize questions and answers. 
%
Attention mechanisms have proven to greatly improve the performance of recurrent networks in many tasks \cite{bahdanau2014ntm, tan2016, rocktaschel2015entailment,rush2015neural,luong2015effective}, and indeed Tan et al.~\cite{tan2016} incorporate a simple attention mechanism in their system.
%
%
%

\section{Preliminaries}
\label{l_sect_prelim}

%
%
\begin{figure*}
\centering
\includegraphics[width=0.65\textwidth]{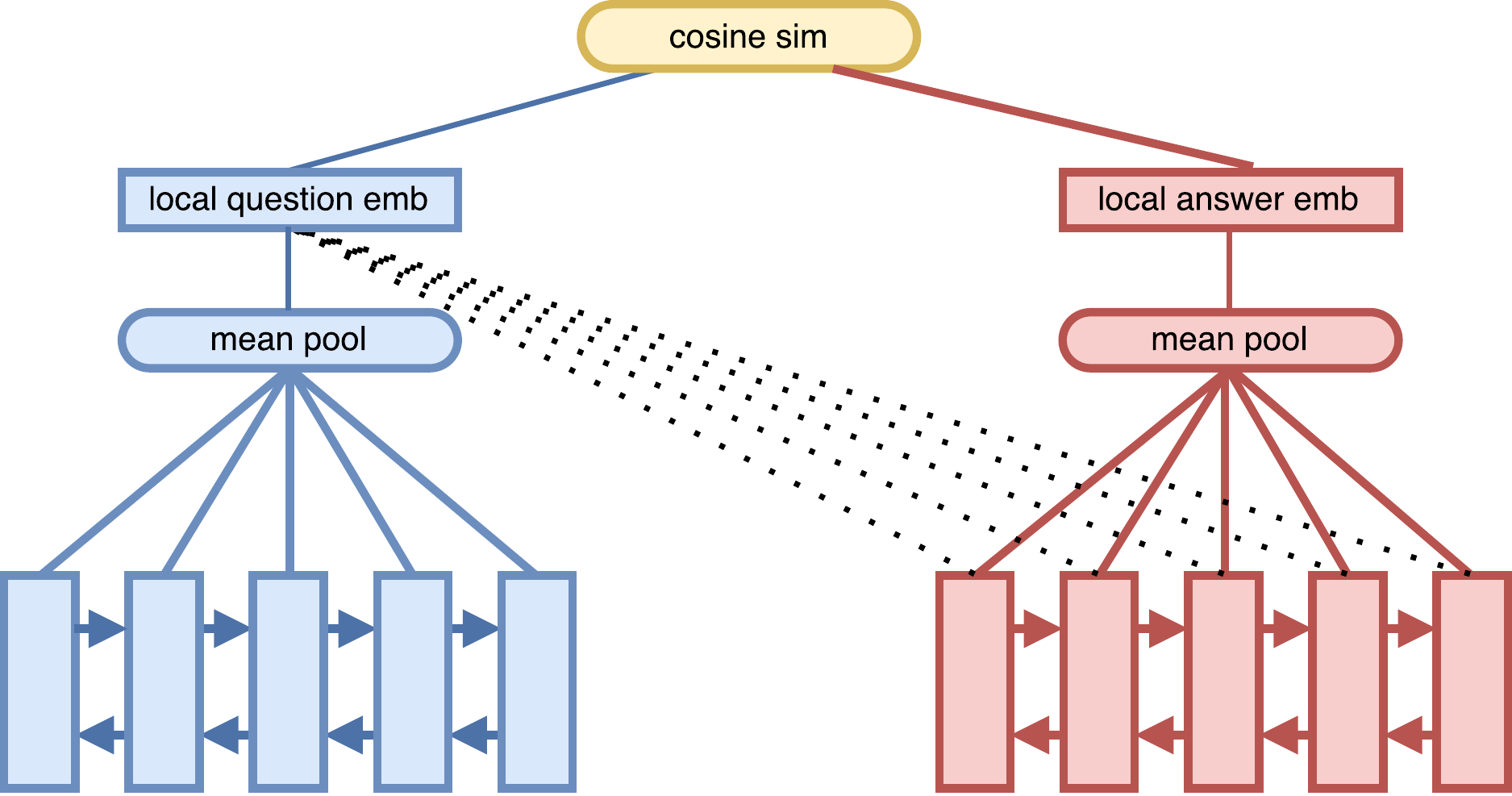}
\caption{Model architecture using answer-localized attention \cite{tan2016}. The left hand side used for the question. The right side of the architecture is used for both the answer and distractor.}
\label{fig:oldModel}
\end{figure*}

\begin{figure*}
\centering
\includegraphics[width=0.8\textwidth]{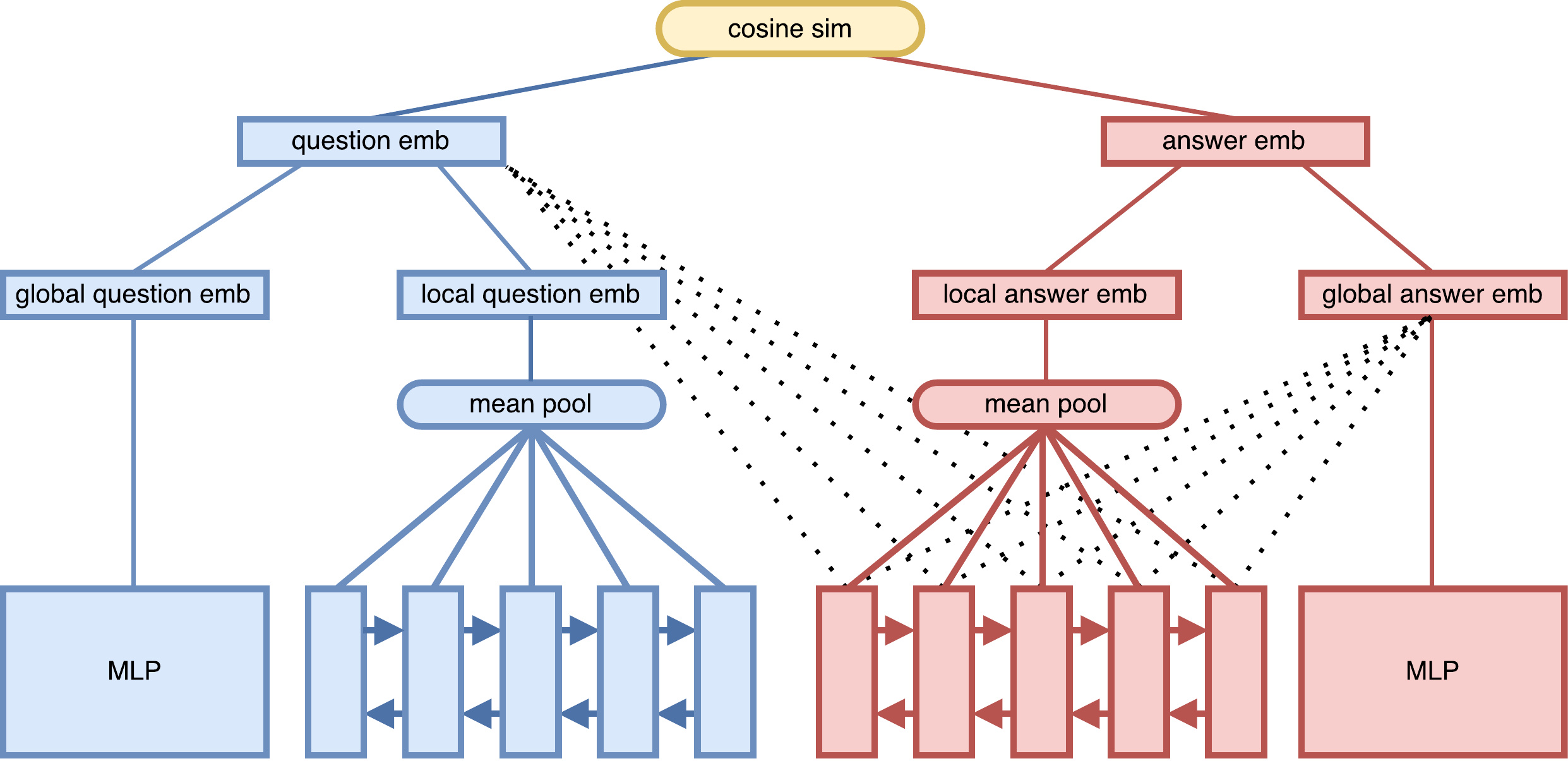}
\caption{Our proposed architecture with augmented attention. As in Figure~\ref{fig:oldModel}, the right side of the model is used to embed answers and distractors.}
\label{fig:newModel}
\end{figure*}

Our approach is similar to the {\em Answer Selection Framework} of Tan et al.~\cite{tan2016}, but we propose a different network architecture and a new attention mechanism. We first provide a high level description of this framework (see the original paper for a more detailed discussion), then discuss our proposed attention mechanism. 

  The framework is based on a neural network with parameters $\theta$ which can embed either a question $q$ or a candidate answer $a$ into low dimensional vectors $r \in \!R^k$. The network can embed a question with no attention, which we denote as $f_{\theta}(q)$, and embed a candidate answer with attention to the question, denoted as $g_{\theta}(a, q)$. We denote the similarity function used as $s(x,y)$ ($s$ may be the dot product function, the cosine similarity function or some other similarity function).

Given a trained network, we compute the similarity between question and answer embeddings:

$$s_i = s(f_{\theta}(q), g_{\theta}(A_i, q))$$
\noindent for any $i \in {1,2,\ldots,k}$ with $A_i$ being the $i$th candidate answer in the pool. We then select the answer yielding the highest similarity $\arg \max_i s_i$. 

The embedding functions, $f_{\theta}$ and $g_{\theta}$, depend on the architecture used and the parameters $\theta$. The network is trained by choosing a loss function $\mathcal{L}$, and using stochastic gradient descent to tune the parameters given the training data. Each training item consists of a question $q$, the correct answer $a^*$ and a distractor $d$ (an incorrect answer). A prominent choice is using a shifted hinge loss, designating that the correct answer must have a higher score than the distractor by at least a certain margin $M$, where the score is based on the similarity to the question. 

$$ \mathcal{L} =\max \Big\{ 0, M - \sigma_{a^*} + \sigma_{d} \Big\} $$ 

where:
$$
\sigma_{a^*} = s \Big(f_{\theta}(q), g_{\theta}(a^*, q) \Big) 
$$ 
$$
\sigma_{d} = s\Big( f_{\theta}(q), g_{\theta}(d, q) \Big)
$$

The above expression has a zero loss if the correct answer has a score higher than the distractor by at least a margin $M$, and the loss linearly increases in the score difference between the correct answer and the distractor. 

Any reasonable neural network design for $f_{\theta}$ can be used to build a working answer-selection systems using the above approach; however, the network design can have a big impact on the system's accuracy. 

\subsection{Embedding Questions and Answers}

Earlier work examined multiple approaches for embedding questions and answers, including convolutional neural networks, recurrent neural networks (RNNs) (sometimes augmented with an attention mechanism) and hybrid designs \cite{fengCNN,tan2016}. 

An RNN design ``digests'' the input sequence, one element at a time, changing its internal state at every timestep. The RNN is based on a cell, a parametrized function mapping a current state and an input element to the new state~\cite{werbos1990backpropagation}. A popular choice for the RNN's cell is the Long Short Term Memory (LSTM) cell~\cite{hochreiter1997long}.

Given a question comprised of words $q=(x_1,x_2,\ldots,x_m)$, we denote the $i$'th output of an LSTM RNN digesting the question as $q_i$; similarly given an answer $a=(y_1,y_2,\ldots,y_n)$ we denote the $j$'th output of an LSTM RNN digesting the question as $a_j$. 

One simple approach is to have the embeddings of the question and answer be the last LSTM output, i.e. $f_{\theta}(q) = q_m$ and $f_{\theta}(a) = a_n$. Note that $q_i,a_i$ are vectors whose dimensionality depends on the dimensionality of the LSTM cell; we denote by $q_{i,j}$ the $j$'th coordinate of the LSTM output at timestep $i$.

Another alternative is to aggregate the LSTM outputs across the different timesteps by taking their coordinate-wise mean (mean-pooling):
$$f_{\theta}(q)_r = \frac{1}{m} \sum_{i=1}^m q_{i,r}$$
\noindent Alternatively, one may aggregate by taking the or coordinate-wise max (max-pooling):
$$f_{\theta}(q)_r = max_{i=1}^m q_{i,r}$$

We use another simple way of embedding the question and answer, which is based on term-frequency (TF) features. Given a vocabulary of words $V=(w_1,\ldots,w_v)$, and a text $p$ we denote the TF representation of $p$ as $p^{\text{tf}} = (d_1,\ldots,d_v)$ where $d_j=1$ if the word $w_j$ occurs in $p$ and otherwise $d_j=0$. \footnote{Another alternative is setting $d_j$ to the {\em number} of times the word $w_j$ appears in $p$. A slightly more complex option is using TF-IDF features~\cite{ramos2003using} or an alternative hand-crafted feature scheme; however we opt for the simpler TF representation, letting the neural network learn how to use the raw information.}

A simple overall embedding of a text $p$ is $p' = W t(p)$ where $W$ is an $v \times d$ matrix, and where $d$ determines the final embedding's dimensionality; the weights of $W$ are typically part of the neural network parameters, to be learned during the training of the network. Instead of a single matrix multiplication, one may use the slightly more elaborate alternative of applying a feedforward network, in order to allow for non-linear embeddings.

We note that a TF representation loses information regarding the {\em order} of the words in the text, but can provide a good global view of key topics discussed in the text.  

Our main contribution is a new design for the neural network that ranks candidate answers for a given question. Our design uses a TF-based representation of the question and answer, and includes a new attention mechanism which uses this global representation when computing the attention weights (in addition to the local information used in existing approaches). We describe existing attention designs (based on local information) in Section~\ref{l_sect_loc_attn}, before proceeding to describe our approach in Section~\ref{l_sect_glob_loc_attn}. 

\subsection{Local Attention}
\label{l_sect_loc_attn}

Early RNN designs were based on applying a deep feedforward network at every timestep, but struggled to cope with longer sequences due to exploding and diminishing gradients \cite{lstm}. Other recurrent cells such as the LSTM and GRU cells \cite{lstm,gru} have been proposed as they alleviate this issue; however, even with such cells, tackling large sequences remains hard~\cite{lstmsSUCK}. Consider using an LSTM to digest a sequence, and taking the final LSTM state to represent the entire sequence; such a design forces the system to represent the entire sequence using a single LSTM state, which is a very narrow channel, making it difficult for the network to represent all the intricacies of a long sequence~\cite{bahdanau2014ntm}. 

Attention mechanisms allow placing varying amounts of emphasis across the entire sequence~\cite{bahdanau2014ntm}, making it easier to process long sequences; in QA, we can give different weights to different parts of the answer while aggregating the LSTM outputs along the different timesteps: 
$$f_{\theta}(a) =  \sum_{i=1}^m \alpha_i a_{i,r}$$
\noindent where $\alpha_i$ denotes the weight (importance) placed on timestep $i$ and $a_{i,r}$ is the $r$th value of the $i$th embedding vector. 

Tan et al.~\cite{tan2016} proposed a very simple attention mechanism for QA, shown in Figure~\ref{fig:oldModel}:
$$ m_{a,q}(i) = W_{ad} a_i + W_{qd} f_{\theta}(q) $$
$$ \alpha_i \propto exp (w_{ms}^T \tanh(m_{a,q}(i))) $$
$$ \hat{a} = \sum_{i=1}^m \alpha_i a_i $$ 
\noindent where $\alpha_i a(i)$ is the weighted hidden layer, $W_{ad}$ and $W_{qd}$ are matrix parameters to be learned, and $w_{ms}$ is a vector parameter to be learned.

\section{Global-Local Attention}
\label{l_sect_glob_loc_attn}

A limitation of the attention mechanism of Tan et al.~\cite{tan2016} is that it only looks at the the embedded question vector and one candidate answer word embedding at a time. Our proposed attention mechanism adds a {\em global} view of the candidate, incorporating information from {\em all} words in the answer. 

\subsection{Creating Global Representations}

One possibility for constructing a global embedding is an RNN design. However, RNN cells tend to focus on the more recent parts of an examined sequence~\cite{lstmsSUCK}. We thus opted for using a term-frequency vector representing the entire answer, as shown in Figure~\ref{fig:newModel}. We denote this representation as:
$$a^{\text{tf}} = (d_1,d_2,\ldots,d_v) $$ 
\noindent where $d_i$ relates to the i'th word in our chosen vocabulary, and $d_i = 1$ if this word appears in the candidate answer, and $d_i = 0$ otherwise. 

Consider a candidate answer $a = (y_1,\ldots,y_n)$, and let $(a_1,\ldots,a_n)$ denote its sequence of RNN LSTM outputs, i.e. $a_i$ denotes the $i$'th output of a RNN LSTM processing this sequence (so $a_i$ is a vector whose dimensionality is as the hidden size of the LSTM cell). We refer to $a_i$ as the local-embedding at time $i$. \footnote{Note that although we call $a_i$ a local embedding, the $i$'th LSTM state does of course take into account other words in the sequence (and not only the $i$'th word). By referring to it as ``local'' we simply mean to say that it is more heavily influenced by the $i$'th word or words close to it in the sequence.}

\subsection{Combining Local and Global Representations to Determine Attention Weights}

The goal of an attention mechanism is to construct an overall representation of the candidate answer $a$, which is later compared to the question representation to determine how well the candidate answers the question; this is achieved by obtaining a set of weights $w_1,\ldots,w_n$ (where $w_i \in \mathbb{R}^+$), and constructing the final answer representation as a weighted average of the LSTM outputs, with these weights. 

Given a candidate answer $a$, we compute the attention coefficient $w_i$ for timestep $i$ as follows.  

First, we combine the local view (the LSTM output, more heavily influenced by the words around timestep $t$) with the global view (based on TF features of all the words in the answer). We begin by taking linear combinations of the TF features then passing them through a $\tanh$ nonlinearity (so that the range of each dimension is bounded in $[-1,1]$):
$$ b^{\text{tf}} = \tanh (W_{1} a^{\text{tf}}) $$
\noindent The weights of the matrix $W_{1}$ are model parameters to be learned, and its dimensions are set so as to map the sparse TF vector $a^{\text{tf}}$ to a dense low dimensional vector (in our implementation $b^{\text{tf}}$ is a 50 dimensional vector). 

Similarly, we take a linear combination of the different dimensions of the local representation $a_i$ (in this case there is no need for the $tanh$ operation, as the LSTM output is already bounded):
$$ b_i^{\text{loc}} = W_{2} a_i $$
\noindent where the weights of the $W_{2}$ are model parameters to be learned (and with dimensions set so that $b_i^{\text{loc}}$ would be a 140 dimensional vector).  

Given a TF representation of a text $x^{\text{tf}}$, whose dimensionality is the size of the vocabulary, and an RNN representation of the text $x^{\text{rnn}}$, with a certain dimentionality $h$, we may wish construct a normalized representation of the text. As the norms of these two parts may differ, simply concatenating these parts may result in a vector dominated by one side. We thus define a joint representation 
$h(x^{\text{tf}}, x^{\text{rnn}})$ as follows. 
%
%

We normalize each part so as to have a desired ratio of norms $\frac{\alpha}{\beta}$ between the RNN and TF representations; this ratio reflects the relative importance of the RNN and TF embeddings in the combined representation (for instance when settings both $\alpha, \beta$ to $1$ both parts would have a unit norm, giving them equal importance): 
$$ c^{\text{tf}}    = \frac{\alpha}{||x^{\text{tf}}||} \cdot x^{\text{tf}} $$
$$ c^{\text{rnn}}   = \frac{\beta} {||x^{\text{rnn}}||} \cdot x^{\text{rnn}} $$ 
\noindent We then concatenate the normalized TF and RNN representations to generate the joint representation:
$$h(x^{\text{tf}}, x^{\text{rnn}}) = c^{\text{tf}} \| c^{\text{rnn}} $$
\noindent where $\|$ represents vector concatenation. 

We construct the local attention representation at the $i$'th word of the answer as:
$$ a_i^{\text{glob-loc}} = h(b^{\text{tf}}, b_i^{\text{loc}}) $$ 
\\ using values of $\alpha=0.5, \beta=1$.

The raw attention coefficient of the $i$'th word in the answer is computed by measuring the similarity of a vector representing the question, and a local-global representation of the answer at word $i$. We build these representations, of matching dimensions, by taking the same number of linear combinations from $a_i^{\text{glob-loc}}$ (the raw global-local representation of the answer at word $i$). Thus the attention weight for the $i$'th word is:

$$
\alpha'_i = sim\Big(  W_{3} a_i^{\text{glob-loc}}, W_{4} f_{\theta}(q)  \Big)
$$
\noindent where $W_2$, $W_3$ are matrices whose weights are parameters to be learned (and whose dimensions are set so that $ W_{3} a_i^{\text{glob-loc}} $ and $W_{4} f_{\theta}(q)$ would be vectors of identical dimensionality, 140 in our implementation), and where $sim$ denotes the cosine similarity between vectors:
$$ sim(u,v) = \frac{u \cdot v}{||u|| \cdot ||v|| } $$ 
\noindent with the $\cdot$ symbol in the nominator denoting the dot product between two vectors.


Finally, we normalize the attention coefficients with respect to their exponent to obtain the final attention weights, by applying the softmax operator on the raw attention coefficients. We take the raw attention coefficients, $\alpha' = (\alpha'_1, \alpha'_2, \ldots, \alpha'_m)$ and define the final attention weights $\alpha = (\alpha_1, \alpha_2, \ldots, \alpha_m)$ where $\alpha_i \propto exp(\alpha'_i)$ and 
$\alpha$ is the result of the softmax operator applied on $\alpha$:
$$ \alpha_i = \frac{\exp{(\alpha'_i)}}{\sum_{j=1}^{m} \exp{(\alpha'_j)}} $$

\subsection{Building the Final Attention Based Representation}

The role of the attention weights is building a final representation of a candidate answer; different answers are ranked based on the similarity of their final representation and a final question representation. 
Similarly to the TF representation of the answer, we denote the TF representation of the question as: $q^{\text{tf}} = (r_1,r_2,\ldots,r_v) $, where $r_i$ relates to the i'th word in our chosen vocabulary, and $r_i = 1$ if this word appears in the question, and $r_i = 0$ otherwise. Our final representation of the question is a joining of the TF representation of the question and the mean pooled RNN question representation (somewhat similarly to how we join the TF and RNN representation when determining the attention weights):
$$ f'_{\theta}(q) = h(q^{\text{tf}}, f_{\theta}(q)) $$ 

Our final representation of the answer is also a joining two parts, a TF part $a^{\text{tf}}$ (as defined earlier) and an attention weighted RNN part $\hat{a}$. We construct $\hat{a}$ as the weighted average of the LSTM outputs, where the weights are the attention weights defined above:
$$ \hat{a} = \sum_{i=1}^m \alpha_i a_i $$ 

The final representation of the answer is thus:
$$ f'_{\theta}(a) = h(a^{\text{tf}}, \hat{a}) $$ 

Figure \ref{fig:newModel} describes the final architecture of our model, showing how we use a TF-based global embedding both in determining the attention weights and in the overall representation of the questions and answers. The dotted lines in the figures indicate that our model's attention weights depend not only on the local embedding but also on the global embedding. 

\subsection{Tuning Parameters to Minimize the Loss}

The loss function $\mathcal{L}$ we use is the shifted hinge loss defined in Section~\ref{l_sect_prelim}. We compute the score of an answer candidate $a$ as the similarity between its final representation $f'_{\theta}(a)$ and the final representation of the question $f'_{\theta}(q)$ 
\footnote{We use the cosine similarity as our similarity function for the loss, though other similarity functions can also be used.} : 
$$sim(f'_{\theta}(q),f'_{\theta}(a))$$ 
\noindent Given the score of the correct answer candidate $\sigma_{a^*} = sim(f'_{\theta}(q),f'_{\theta}(a))$ and the score of a distractor (incorrect) candidate $d$, $\sigma_d = sim(f'_{\theta}(q),f'_{\theta}(d))$, our loss is 
$\mathcal{L} =\max \Big\{ 0, M - \sigma_{a^*} + \sigma_{d} \Big\}$. 

The above loss relates to a single training item (consisting of a single question, its correct answer and an incorrect candidate answer). Training the neural network parameters involves iteratively examining items in a dataset consisting of many training items (each containing a question, its correct answer and a distractor) and modifying the current network parameters. We train our system using variant of stochastic gradient descent (SGD) with the Adam optimization~\cite{kingma2014adam}.

\section{Empirical Evaluation}

We evaluate our proposed neural network design in a similar manner to earlier evaluations of Siamese neural network designs~\cite{yang2015wikiqa,severyn2015learning}, where a neural network is trained to embed both questions and candidate answers as low dimensional vectors. 

\subsection{Experiment Setup} \label{sec:setup}

\begin{figure*}[h!t]
\includegraphics[width=\textwidth]{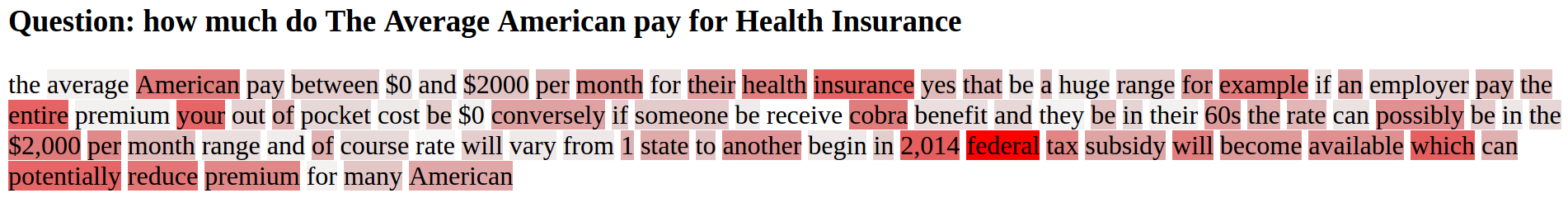}
\includegraphics[width=0.8\textwidth]{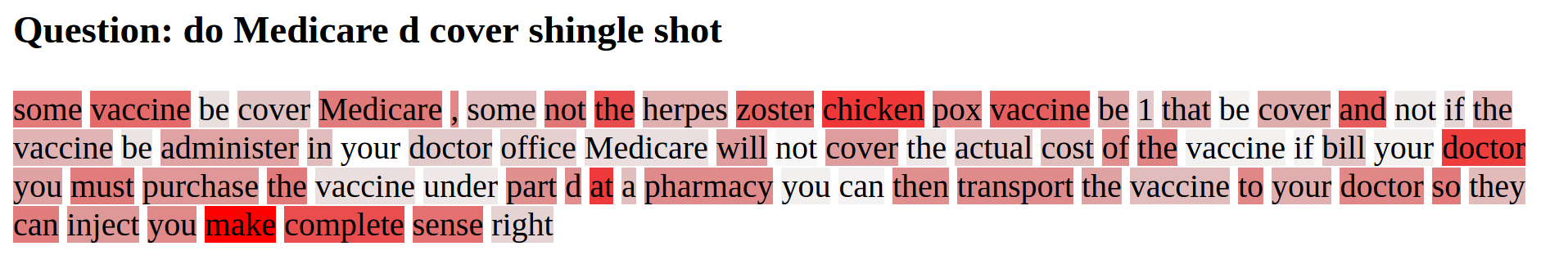}
\caption{A visualization of the attention weights for each word in a correct answer to a question. These examples show how the attention mechanism is focusing on relevant parts of the correct answer (although the attention is still quite noisy).}
\label{fig-example-attn-weights}
\end{figure*}

\begin{figure}[h!t]
\centering
\includegraphics[width=0.5\textwidth]{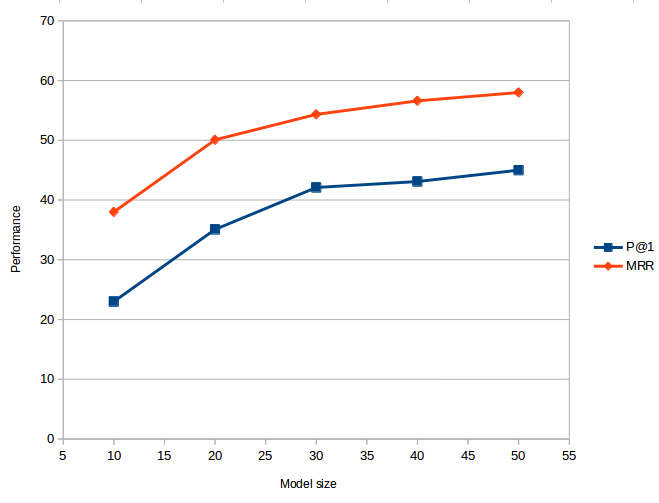}
\caption{Performance of our system on InsuranceQA for various model sizes $h$ (both the LSTM hidden layer size and embedding size)}
\label{fig-size-to-perf}
\end{figure}
%
%
%

We use the InsuranceQA dataset and its  evaluation framework~\cite{fengCNN}. 
The InsuranceQA dataset contains question and answer pairs from the insurance domain, with roughly 25,000 unique answers, and is already partitioned into a training set and two test sets, called test 1 and test 2. 

The InsuranceQA dataset has relatively short questions (mean length of 7). However, the answers are typically very long (mean length of 94). 

At test time the system takes as input a question $q$ and a pool of candidate answers $P=(a_1,a_2,\ldots,a_k)$ and is asked to select the best matching answer $a^*$ to the question from the pool. The InsuranceQA comes with answer pools of size $k=500$, consisting of the correct answers and random distractors chosen from the set of answers to other questions. 

State-of-the-art results for InsuranceQA were achieved by Tan et al~\cite{tan2016}, which also provide a comparison with several baselines: Bag-of-words (with IDF weighted sum of word vectors and cosine similarity based ranking), the Metzler-Bendersky IR model~\cite{bendersky2010}, and ~\cite{fengCNN} - the CNN based Architecture-II and Architecture-II with Geometricmean of Euclidean and Sigmoid Dot product (GESD).

We implemented our model in TensorFlow~\cite{abadi2016tensorflow} and conducted experiments on our GPU cluster. 

We use the same hidden layer sizes and embedding size as Tan et al.~\cite{tan2016}: $h=141$ for the bidirectional LSTM size and an embedding size of $e=100$; this allows us to investigate the impact of our proposed attention mechanism. \footnote{As is the case with many neural networks, increasing the hidden layer size or embedding size can improve the performance on our InsuranceQA models; we compare our performance to the work of Tan et al.~\cite{tan2016} with the same hidden and embedding sizes; similarly to them we use embeddings pre-trained using Word2Vec~\cite{mikolov2013} and avoid overfitting by applying early stopping (we also apply Dropout~\cite{dropout,zaremba2014recurrent}). } 

\begin{table}[h!t]
\small
\centering
  \begin{tabular}{ | l | c | c | }
    \hline
    Model & Test1 & Test2 \\ 
    \hline
    \hline
    Bag-of-words & 32.1 & 32.2 \\ 
    \hline
    Metzler-Bendersky & 55.1 & 50.8 \\ 
    \hline
    Arch-II~\cite{fengCNN} & 62.8 & 59.2 \\ 
    \hline
    Arch-II GSED~\cite{fengCNN} & 65.3 &   61.0 \\ 
    \hline 
    Attention LSTM~\cite{tan2016} & 69.0 & 64.8 \\  
    \hline
    \hline
    TF-LSTM Concatenation & 62.1 & {61.5} \\ 
    \hline    
    Local-Global Attention & {\bf 70.1} & {\bf 67.4} \\ 
    \hline    
  \end{tabular}
  \vspace{0.2cm}
\caption{Performance of various models on InsuranceQA}
\label{table:perf_insqa}
\end{table}

\subsection{Results}

Table~\ref{table:perf_insqa} presents the results of our model and the various baselines for InsuranceQA. The performance metric used here is P@1, the proportion of instances where a correct answer was ranked higher than all other distractors in the pool. The table shows that our model outperforms the previous baselines.  

We have also examined the performance of our model as a function of its size (determining the system's runtime and memory consumption). We used different values $h \in \{10,20,30,40,50\}$ for both the size of the LSTM's hidden layer size and embedding size, and examined the performance of the resulting QA system on InsuranceQA. Our results are given in Figure~\ref{fig-size-to-perf}, which shows both the P@1 metric and the mean reciprocal rank (MRR)~\cite{craswell2009mean,chapelle2009expected} \footnote{The MRR metric assigns the model partial credit even in cases where the highest ranking candidate is an incorrect answer, with the score depending on the highest rank of a correct answer. }

Figure~\ref{fig-size-to-perf} shows that performance improves as the model gets larger, but the returns on extending the model size quickly diminish. Interestingly, even relatively small models achieve a reasonable question answering performance. 

To show our attention mechanism is necessary to achieve good performance, we also construct a model that simply concatenates the output of the feedforward network (on TF features) and the output of the bidirectional LSTM, called TF-LSTM concatenation. While this model does make use of TF-based features in addition to the LSTM state of the RNN, it does not use an attention mechanism to allow it to focus on the more relevant parts of the text. As the table shows, the performance of the TF-LSTM model is significantly lower than that of our model with the global-local attention mechanism. This indicates that the improved performance stems from the model's improved ability to focus on the relevant parts of the answer (and not simply from having a larger capacity and including TF-features).

Finally, we examine the the attention model's weights to evaluate it qualitatively. Figure~\ref{fig-example-attn-weights} visualizes the weights for two question-answer pairs, where the color intensity reflects the relative weight placed on the word (the $\alpha_i$ coefficients discussed earlier). The figure shows that our attention model can focus on the parts of the candidate answer that are most relevant for the given question. 



\section{Conclusion}

We proposed a new neural design for answer selection, using 
an augmented attention mechanism, which combines both local and global information when determining the attention weight to place at a given timestep. Our analysis shows that our design outperforms earlier designs based on a simpler attention mechanism which only considers the local view. 

Several questions remain open for future research. First, the TF-based global view of our design was extremely simple; could a more elaborate design, possibly using convolutional neural networks, achieve better performance? 

Second, our attention mechanism joins the local and global information in a very simple manner, by normalizing each vector and concatenating the normalized vectors. Could a more sophisticated joining of this information, perhaps allowing for more interaction between the parts, help further improve the performance of our mechanism?

Finally, can the underlying principles of our global-local attention design improve the performance of other systems, such as machine translation or image processing systems? 

\bibliographystyle{IEEEtran}
\bibliography{attn}

\begin{thebibliography}{10}
\providecommand{\url}[1]{#1}
\csname url@samestyle\endcsname
\providecommand{\newblock}{\relax}
\providecommand{\bibinfo}[2]{#2}
\providecommand{\BIBentrySTDinterwordspacing}{\spaceskip=0pt\relax}
\providecommand{\BIBentryALTinterwordstretchfactor}{4}
\providecommand{\BIBentryALTinterwordspacing}{\spaceskip=\fontdimen2\font plus
\BIBentryALTinterwordstretchfactor\fontdimen3\font minus
  \fontdimen4\font\relax}
\providecommand{\BIBforeignlanguage}[2]{{%
\expandafter\ifx\csname l@#1\endcsname\relax
\typeout{** WARNING: IEEEtran.bst: No hyphenation pattern has been}%
\typeout{** loaded for the language `#1'. Using the pattern for}%
\typeout{** the default language instead.}%
\else
\language=\csname l@#1\endcsname
\fi
#2}}
\providecommand{\BIBdecl}{\relax}
\BIBdecl

\bibitem{burke1997question}
R.~D. Burke, K.~J. Hammond, V.~Kulyukin, S.~L. Lytinen, N.~Tomuro, and
  S.~Schoenberg, ``Question answering from frequently asked question files:
  Experiences with the faq finder system,'' \emph{AI magazine}, vol.~18, no.~2,
  p.~57, 1997.

\bibitem{voorhees1999trec}
E.~M. Voorhees \emph{et~al.}, ``The trec-8 question answering track report.''
  in \emph{Trec}, vol.~99, 1999, pp. 77--82.

\bibitem{kwok2001scaling}
C.~Kwok, O.~Etzioni, and D.~S. Weld, ``Scaling question answering to the web,''
  \emph{ACM Transactions on Information Systems (TOIS)}, vol.~19, no.~3, pp.
  242--262, 2001.

\bibitem{hirschman2001natural}
L.~Hirschman and R.~Gaizauskas, ``Natural language question answering: the view
  from here,'' \emph{natural language engineering}, vol.~7, no.~4, pp.
  275--300, 2001.

\bibitem{ravichandran2002learning}
D.~Ravichandran and E.~Hovy, ``Learning surface text patterns for a question
  answering system,'' in \emph{Proceedings of the 40th annual meeting on
  association for computational linguistics}.\hskip 1em plus 0.5em minus
  0.4em\relax Association for Computational Linguistics, 2002, pp. 41--47.

\bibitem{xu2002trec}
J.~Xu, A.~Licuanan, J.~May, S.~Miller, and R.~M. Weischedel, ``Trec 2002 qa at
  bbn: Answer selection and confidence estimation.'' in \emph{TREC}, vol.~54,
  2002, p.~90.

\bibitem{jijkoun2004answer}
V.~Jijkoun and M.~De~Rijke, ``Answer selection in a multi-stream open domain
  question answering system,'' in \emph{ECIR}, vol. 2997.\hskip 1em plus 0.5em
  minus 0.4em\relax Springer, 2004, pp. 99--111.

\bibitem{ko2007probabilistic}
J.~Ko, L.~Si, and E.~Nyberg, ``A probabilistic framework for answer selection
  in question answering.'' in \emph{HLT-NAACL}, 2007, pp. 524--531.

\bibitem{lee2009model}
C.~T. Lee, E.~M. Rodrigues, G.~Kazai, N.~Milic-Frayling, and A.~Ignjatovic,
  ``Model for voter scoring and best answer selection in community q\&a
  services,'' in \emph{Web Intelligence and Intelligent Agent Technologies,
  2009. WI-IAT'09. IEEE/WIC/ACM International Joint Conferences on},
  vol.~1.\hskip 1em plus 0.5em minus 0.4em\relax IEEE, 2009, pp. 116--123.

\bibitem{severyn2013automatic}
A.~Severyn and A.~Moschitti, ``Automatic feature engineering for answer
  selection and extraction.'' in \emph{EMNLP}, vol.~13, 2013, pp. 458--467.

\bibitem{kolomiyets2011survey}
O.~Kolomiyets and M.-F. Moens, ``A survey on question answering technology from
  an information retrieval perspective,'' \emph{Information Sciences}, vol.
  181, no.~24, pp. 5412--5434, 2011.

\bibitem{allam2012question}
A.~M.~N. Allam and M.~H. Haggag, ``The question answering systems: A survey,''
  \emph{International Journal of Research and Reviews in Information Sciences
  (IJRRIS)}, vol.~2, no.~3, 2012.

\bibitem{clarke2001exploiting}
C.~L. Clarke, G.~V. Cormack, and T.~R. Lynam, ``Exploiting redundancy in
  question answering,'' in \emph{Proceedings of the 24th annual international
  ACM SIGIR conference on Research and development in information
  retrieval}.\hskip 1em plus 0.5em minus 0.4em\relax ACM, 2001, pp. 358--365.

\bibitem{parsetreeManning}
M.~Wang and C.~D. Manning, ``Probabilistic tree-edit models with structured
  latent variables for textual entailment and question answering,'' in
  \emph{Proceedings of the 23rd International Conference on Computational
  Linguistics}.\hskip 1em plus 0.5em minus 0.4em\relax Association for
  Computational Linguistics, 2010, pp. 1164--1172.

\bibitem{wang2007jeopardy}
M.~Wang, N.~A. Smith, and T.~Mitamura, ``What is the jeopardy model? a
  quasi-synchronous grammar for qa.'' in \emph{EMNLP-CoNLL}, vol.~7, 2007, pp.
  22--32.

\bibitem{krizhevsky2012imagenet}
A.~Krizhevsky, I.~Sutskever, and G.~E. Hinton, ``Imagenet classification with
  deep convolutional neural networks,'' in \emph{Advances in neural information
  processing systems}, 2012, pp. 1097--1105.

\bibitem{zhou2014learning}
B.~Zhou, A.~Lapedriza, J.~Xiao, A.~Torralba, and A.~Oliva, ``Learning deep
  features for scene recognition using places database,'' in \emph{Advances in
  neural information processing systems}, 2014, pp. 487--495.

\bibitem{lewenberg2016predicting}
Y.~Lewenberg, Y.~Bachrach, S.~Shankar, and A.~Criminisi, ``Predicting personal
  traits from facial images using convolutional neural networks augmented with
  facial landmark information.'' in \emph{IJCAI}, 2016, pp. 1676--1682.

\bibitem{albarqouni2016aggnet}
S.~Albarqouni, C.~Baur, F.~Achilles, V.~Belagiannis, S.~Demirci, and N.~Navab,
  ``Aggnet: deep learning from crowds for mitosis detection in breast cancer
  histology images,'' \emph{IEEE transactions on medical imaging}, vol.~35,
  no.~5, pp. 1313--1321, 2016.

\bibitem{gaunt2016training}
A.~Gaunt, D.~Borsa, and Y.~Bachrach, ``Training deep neural nets to aggregate
  crowdsourced responses,'' in \emph{Proceedings of the Thirty-Second
  Conference on Uncertainty in Artificial Intelligence. AUAI Press}, 2016, p.
  242251.

\bibitem{graves2013speech}
A.~Graves, A.-r. Mohamed, and G.~Hinton, ``Speech recognition with deep
  recurrent neural networks,'' in \emph{Acoustics, speech and signal processing
  (icassp), 2013 ieee international conference on}.\hskip 1em plus 0.5em minus
  0.4em\relax IEEE, 2013, pp. 6645--6649.

\bibitem{bahdanau2014ntm}
D.~Bahdanau, K.~Cho, and Y.~Bengio, ``Neural machine translation by jointly
  learning to align and translate,'' \emph{arXiv preprint arXiv:1409.0473},
  2014.

\bibitem{li2015diversity}
J.~Li, M.~Galley, C.~Brockett, J.~Gao, and B.~Dolan, ``A diversity-promoting
  objective function for neural conversation models,'' \emph{arXiv preprint
  arXiv:1510.03055}, 2015.

\bibitem{kandasamy2017batch}
K.~Kandasamy, Y.~Bachrach, R.~Tomioka, D.~Tarlow, and D.~Carter, ``Batch policy
  gradient methods for improving neural conversation models,'' \emph{arXiv
  preprint arXiv:1702.03334}, 2017.

\bibitem{shao2017generating}
L.~Shao, S.~Gouws, D.~Britz, A.~Goldie, B.~Strope, and R.~Kurzweil,
  ``Generating long and diverse responses with neural conversation models,''
  \emph{arXiv preprint arXiv:1701.03185}, 2017.

\bibitem{fengCNN}
M.~Feng, B.~Xiang, M.~R. Glass, L.~Wang, and B.~Zhou, ``Applying deep learning
  to answer selection: A study and an open task,'' in \emph{Automatic Speech
  Recognition and Understanding (ASRU), 2015 IEEE Workshop on}.\hskip 1em plus
  0.5em minus 0.4em\relax IEEE, 2015, pp. 813--820.

\bibitem{tan2016}
M.~Tan, C.~dos Santos, B.~Xiang, and B.~Zhou, ``Improved representation
  learning for question answer matching,'' in \emph{Proceedings of the 54th
  Annual Meeting of the Association for Computational Linguistics}, 2016.

\bibitem{weston2015towards}
J.~Weston, A.~Bordes, S.~Chopra, A.~M. Rush, B.~van Merri{\"e}nboer, A.~Joulin,
  and T.~Mikolov, ``Towards ai-complete question answering: A set of
  prerequisite toy tasks,'' \emph{arXiv preprint arXiv:1502.05698}, 2015.

\bibitem{rajpurkar2016squad}
P.~Rajpurkar, J.~Zhang, K.~Lopyrev, and P.~Liang, ``Squad: 100,000+ questions
  for machine comprehension of text,'' \emph{arXiv preprint arXiv:1606.05250},
  2016.

\bibitem{yang2015wikiqa}
Y.~Yang, W.-t. Yih, and C.~Meek, ``Wikiqa: A challenge dataset for open-domain
  question answering.'' in \emph{EMNLP}.\hskip 1em plus 0.5em minus 0.4em\relax
  Citeseer, 2015, pp. 2013--2018.

\bibitem{ChangWordnet}
W.-t. Y. M.-W. Chang and C.~M.~A. Pastusiak, ``Question answering using
  enhanced lexical semantic models,'' 2013.

\bibitem{mueller2016siamese}
J.~Mueller and A.~Thyagarajan, ``Siamese recurrent architectures for learning
  sentence similarity.'' in \emph{AAAI}, 2016, pp. 2786--2792.

\bibitem{rocktaschel2015entailment}
T.~Rockt{\"a}schel, E.~Grefenstette, K.~M. Hermann, T.~Ko{\v{c}}isk{\`y}, and
  P.~Blunsom, ``Reasoning about entailment with neural attention,'' \emph{arXiv
  preprint arXiv:1509.06664}, 2015.

\bibitem{rush2015neural}
A.~M. Rush, S.~Chopra, and J.~Weston, ``A neural attention model for
  abstractive sentence summarization,'' \emph{arXiv preprint arXiv:1509.00685},
  2015.

\bibitem{luong2015effective}
M.-T. Luong, H.~Pham, and C.~D. Manning, ``Effective approaches to
  attention-based neural machine translation,'' \emph{arXiv preprint
  arXiv:1508.04025}, 2015.

\bibitem{werbos1990backpropagation}
P.~J. Werbos, ``Backpropagation through time: what it does and how to do it,''
  \emph{Proceedings of the IEEE}, vol.~78, no.~10, pp. 1550--1560, 1990.

\bibitem{hochreiter1997long}
S.~Hochreiter and J.~Schmidhuber, ``Long short-term memory,'' \emph{Neural
  computation}, vol.~9, no.~8, pp. 1735--1780, 1997.

\bibitem{ramos2003using}
J.~Ramos \emph{et~al.}, ``Using tf-idf to determine word relevance in document
  queries,'' in \emph{Proceedings of the first instructional conference on
  machine learning}, vol. 242, 2003, pp. 133--142.

\bibitem{lstm}
S.~Hochreiter and J.~Schmidhuber, ``Long short-term memory,'' \emph{Neural
  computation}, vol.~9, no.~8, pp. 1735--1780, 1997.

\bibitem{gru}
\BIBentryALTinterwordspacing
J.~Chung, {\c{C}}.~G{\"{u}}l{\c{c}}ehre, K.~Cho, and Y.~Bengio, ``Empirical
  evaluation of gated recurrent neural networks on sequence modeling,''
  \emph{CoRR}, vol. abs/1412.3555, 2014. [Online]. Available:
  \url{http://arxiv.org/abs/1412.3555}
\BIBentrySTDinterwordspacing

\bibitem{lstmsSUCK}
\BIBentryALTinterwordspacing
T.~Mikolov, A.~Joulin, S.~Chopra, M.~Mathieu, and M.~Ranzato, ``Learning longer
  memory in recurrent neural networks,'' \emph{CoRR}, vol. abs/1412.7753, 2014.
  [Online]. Available: \url{http://arxiv.org/abs/1412.7753}
\BIBentrySTDinterwordspacing

\bibitem{kingma2014adam}
D.~Kingma and J.~Ba, ``Adam: A method for stochastic optimization,''
  \emph{arXiv preprint arXiv:1412.6980}, 2014.

\bibitem{severyn2015learning}
A.~Severyn and A.~Moschitti, ``Learning to rank short text pairs with
  convolutional deep neural networks,'' in \emph{Proceedings of the 38th
  International ACM SIGIR Conference on Research and Development in Information
  Retrieval}.\hskip 1em plus 0.5em minus 0.4em\relax ACM, 2015, pp. 373--382.

\bibitem{bendersky2010}
M.~Bendersky, D.~Metzler, and W.~B. Croft, ``Learning concept importance using
  a weighted dependence model,'' in \emph{Proceedings of the third ACM
  international conference on Web search and data mining}.\hskip 1em plus 0.5em
  minus 0.4em\relax ACM, 2010, pp. 31--40.

\bibitem{abadi2016tensorflow}
M.~Abadi, A.~Agarwal, P.~Barham, E.~Brevdo, Z.~Chen, C.~Citro, G.~S. Corrado,
  A.~Davis, J.~Dean, M.~Devin \emph{et~al.}, ``Tensorflow: Large-scale machine
  learning on heterogeneous distributed systems,'' \emph{arXiv preprint
  arXiv:1603.04467}, 2016.

\bibitem{mikolov2013}
T.~Mikolov, I.~Sutskever, K.~Chen, G.~S. Corrado, and J.~Dean, ``Distributed
  representations of words and phrases and their compositionality,'' in
  \emph{Advances in neural information processing systems}, 2013, pp.
  3111--3119.

\bibitem{dropout}
N.~Srivastava, G.~E. Hinton, A.~Krizhevsky, I.~Sutskever, and R.~Salakhutdinov,
  ``Dropout: a simple way to prevent neural networks from overfitting.''
  \emph{Journal of Machine Learning Research}, vol.~15, no.~1, pp. 1929--1958,
  2014.

\bibitem{zaremba2014recurrent}
W.~Zaremba, I.~Sutskever, and O.~Vinyals, ``Recurrent neural network
  regularization,'' \emph{arXiv preprint arXiv:1409.2329}, 2014.

\bibitem{craswell2009mean}
N.~Craswell, ``Mean reciprocal rank,'' in \emph{Encyclopedia of Database
  Systems}.\hskip 1em plus 0.5em minus 0.4em\relax Springer, 2009, pp.
  1703--1703.

\bibitem{chapelle2009expected}
O.~Chapelle, D.~Metlzer, Y.~Zhang, and P.~Grinspan, ``Expected reciprocal rank
  for graded relevance,'' in \emph{Proceedings of the 18th ACM conference on
  Information and knowledge management}.\hskip 1em plus 0.5em minus 0.4em\relax
  ACM, 2009, pp. 621--630.

\end{thebibliography}

\end{document}